# Deep Learning Aided Vision System for Planetary Rovers

Lomash Relia
Department of Computer Engineering
Devang Patel Institute of Advanced Technology and Research
Anand, India
lomashrelia@gmail.com

Jai G Singla
Space Applications Centre
Indian Space Research Organization
Ahmedabad, India
jaisingla@sac.isro.gov.in

Amitabh
Space Applications Centre
Indian Space Research Organization
Ahmedabad, India
amitabh@sac.isro.gov.in

Nitant Dube
Space Applications Centre
Indian Space Research Organization
Ahmedabad, India
nitant@sac.isro.gov.in

*Abstract*— **This study presents a vision system for planetary rovers, combining real-time perception with offline terrain reconstruction. The real-time module integrates CLAHE-enhanced stereo imagery, YOLOv11n-based object detection, and a neural network to estimate object distances. The offline module uses the Depth Anything V2 (metric) monocular depth estimation model to generate depth maps from captured images, which are fused into dense point clouds using Open3D. Real-world distance estimates from the real-time pipeline provide reliable metric context alongside the qualitative reconstructions. Evaluation on Chandrayaan-3 NavCam stereo imagery, benchmarked against a CAHV-based utility, shows that the neural network achieves a median depth error of 2.26 cm within a 1–10 meter range. The object detection model maintains a balanced precision–recall trade-off on grayscale lunar scenes. This architecture offers a scalable, compute-efficient vision solution for autonomous planetary exploration.**

## I. Introduction

Autonomous navigation presents a significant challenge for planetary exploration missions, where communication delays and unstructured extraterrestrial environments necessitate real-time perception and decision-making capabilities. Planetary rovers must be capable of reliably detecting and classifying obstacles, assessing terrain traversability, and supporting scientific operations with minimal ground intervention. Among the sensing modalities available, stereo vision has emerged as a preferred solution, offering passive depth perception with lower power requirements compared to active sensors such as LiDAR.

The successful deployment of stereo vision systems in missions such as NASA's Mars Exploration Rovers—Sojourner, Curiosity, and Perseverance [1]—as well as ISRO's Chandrayaan-3 [2], has demonstrated the maturity of this technology for hazard detection and terrain mapping [3]. However, the computational limitations of rover hardware, combined with the sparse visual features of lunar and Martian terrains, continue to constrain the effectiveness of real-time, high-fidelity stereo processing.

Classical geometric approaches such as CAHVOR modelling with least-squares triangulation deliver high depth accuracy [3] but are computationally intensive, making them impractical for low-power onboard execution. Similarly, dense stereo matching algorithms like Semi-Global Matching (SGM) [4] degrade in performance on low-texture regolith surfaces. Recent machine learning-based approaches [5], [6], [7], [8] offer promising alternatives but remain largely challenging for real-time use in harsh extraterrestrial environments due to their power and memory requirements.

We propose a stereo vision framework tailored to the constraints of planetary rovers. The real-time module performs lightweight object detection using YOLO11n [9] and distance estimation via a neural network trained on stereo keypoints. In parallel, a secondary offline pipeline leverages monocular depth estimation for high-resolution terrain reconstruction using captured imagery. The system is evaluated on Chandrayaan-3 NavCam stereo imagery, demonstrating both computational efficiency and metric accuracy suitable for autonomous navigation. By decoupling immediate perception from post-mission analysis, the proposed framework enables reliable obstacle detection in the field while supporting passive generation of high-quality 3D reconstructions for scientific applications.

## II. Related Work

Stereo vision has played a foundational role in planetary rover autonomy since NASA's Mars Exploration Rovers (MER), which employed stereo cameras for real-time hazard detection and visual odometry [5]. Subsequent missions such as Curiosity and Perseverance advanced these capabilities by enabling enhanced 3D terrain mapping to support long-range navigation. More recently, stereo navigation camera (NavCam) imagery from ISRO's Chandrayaan-3 Pragyan rover has been used to generate digital elevation models (DEMs) for lunar surface analysis [2].

Classical approaches to 3D reconstruction commonly rely on CAHV or CAHVOR camera models. Yakimovsky and Cunningham [10] first introduced CAHV modelling for stereo vision in space applications, using least-squares triangulation to compute 3D coordinates. This involves solving an overdetermined system of linear equations for each matched feature pair from the stereo images—specifically using their pixel coordinates—to estimate the real-world X, Y and Z positions with respect to origin of the defined world coordinate system. The CAHVOR extension accounts for lens distortion and improves modelling accuracy, but remains computationally expensive for large-scale keypoint matching. Feature-based correspondence methods such as SIFT offer robust matching under geometric and photometric transformations [11]. Dense stereo matching algorithms like Semi-Global Matching (SGM) [4] offer full-field depth estimation but suffer from reduced reliability in low-texture terrains like lunar regolith especially if the images are monochrome.

To address these limitations, recent studies have explored machine learning-based alternatives. Lightweight neural networks can approximate geometric triangulation and reduce computational overhead by operating on input features in batch mode. Object detection models like YOLO11 Nano [9] have shown effectiveness in performing fast hazard recognition with limited compute requirements. Likewise, monocular depth estimation models such as Depth Anything V2 [12] offer high-resolution reconstructions from single grayscale images.

This work bridges the gap by introducing a vision pipeline that leverages both paradigms: efficient, real-time object detection and depth estimation using deep learning, coupled with offline monocular 3D reconstruction for comprehensive terrain construction. The approach is validated on Chandrayaan-3 NavCam stereo imagery, demonstrating its feasibility for use in resource-constrained planetary missions.

## III. Methodology

The Chandrayaan-3 Pragyan rover was equipped with a calibrated stereo camera system that captured monochrome image pairs with a fixed baseline of 0.24 meters, facilitating accurate depth perception in the lunar environment. The stereo setup featured a 39-degree field of view, enabling coverage of terrain with significant illumination variability caused by lunar shadows. As shown in Figure 1, Contrast Limited Adaptive Histogram Equalization (CLAHE) was applied to enhance local contrast, using a clip limit of 2.0 and a tile grid size of 8×8. This preprocessing step significantly improved feature visibility in low-texture regolith regions, thereby enhancing the robustness of subsequent detection and matching algorithms.

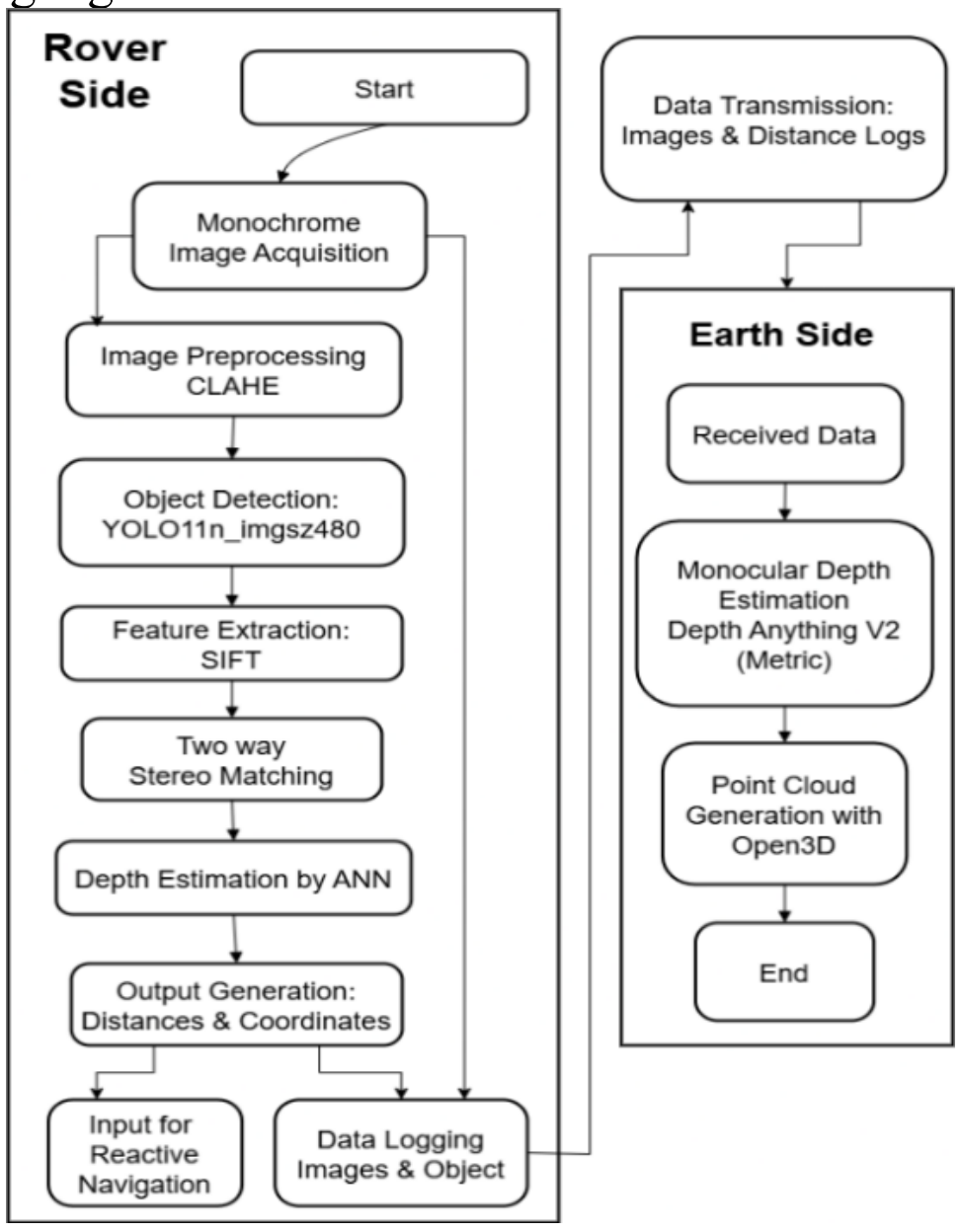

*Figure 1 Pipeline Flowchart*

Figure 1 illustrates the overall flowchart of the proposed dual-pipeline architecture. On the rover side, images undergo CLAHE-based preprocessing and are passed to the YOLO11n model for object detection. SIFT features are extracted within detected bounding boxes only and matched bidirectionally to identify corresponding objects. A neural network then estimates the 3D position (X, Y, Z) from the matched features but we only utilize the distance value as the 2D object positions are already defined by the bounding box coordinates. Objects lacking at least 4 correspondences due to illumination differences or limited overlap are excluded to maintain triangulation reliability.

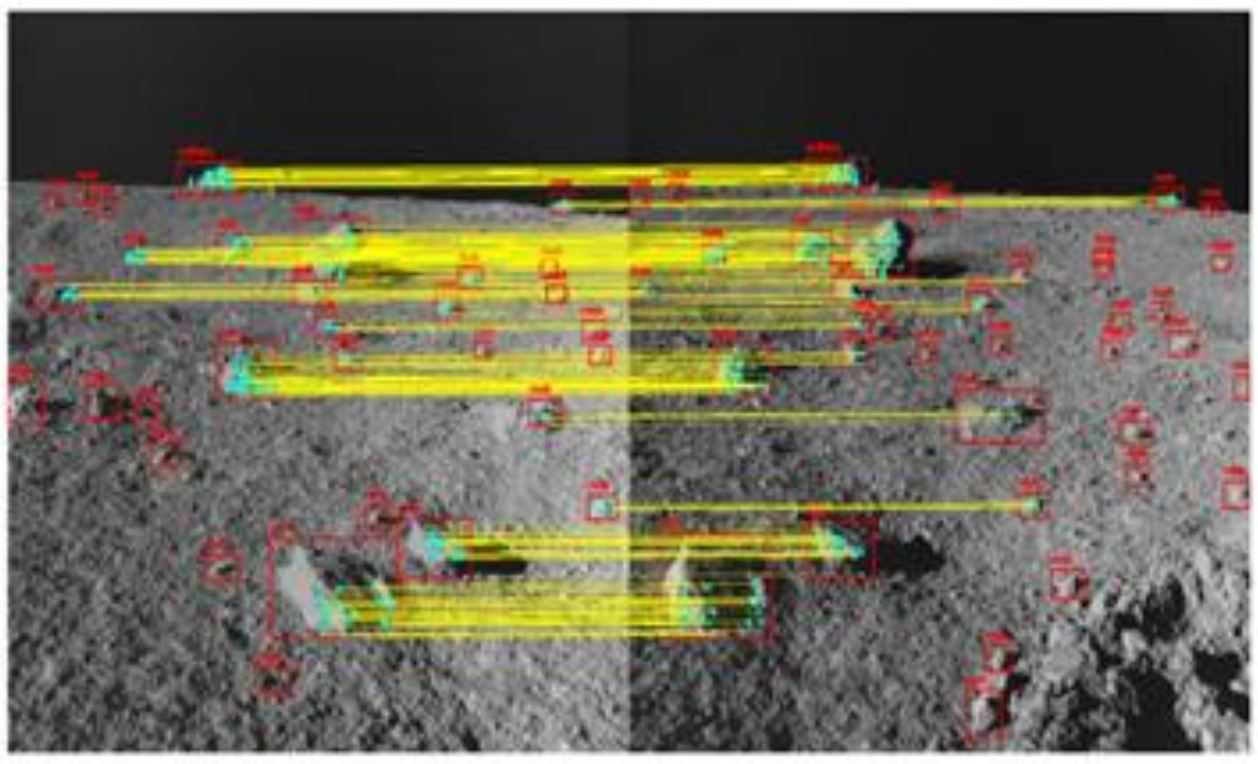

| ID | class | median distance | bbox | left coordinates |
|---|---|---|---|---|
| 1 | rock | 140.55 | [341.32, 451.81, 435.37, 532.89] | [410.14, 519.4] |
| 7 | rock | 150.18 | [539.34, 456.86, 583.08, 494.5] | [563.05, 488.61] |
| 6 | rock | 204.1 | [704.77, 392.36, 742.36, 422.62] | [725.35, 415.79] |
| 4 | rock | 214.63 | [470.27, 371.35, 518.99, 408.24] | [505.19, 395.45] |
| 2 | rock | 228.22 | [479.11, 289.63, 598.28, 372.94] | [562.92, 313.59] |
| 9 | rock | 336.99 | [93.39, 377.77, 122.44, 397.98] | [120.05, 390.77] |
| 0 | rock | 349.33 | [346.96, 307.84, 402.03, 350.56] | [378.67, 324.98] |
| 3 | artifact | 950 | [281.13, 208.25, 380.87, 254.53] | [349.92, 238.2] |

*Figure 2 Example detections and output table*

The output, as shown in Figure 2, includes object class, bounding box, and median distance per object, which is logged and transmitted along with images for Earth-side analysis. Offline, Depth Anything V2 generates monocular depth maps from the captured left images. Open3D fuses this output with the previously estimated 3D points to reconstruct a dense point cloud of the lunar surface.

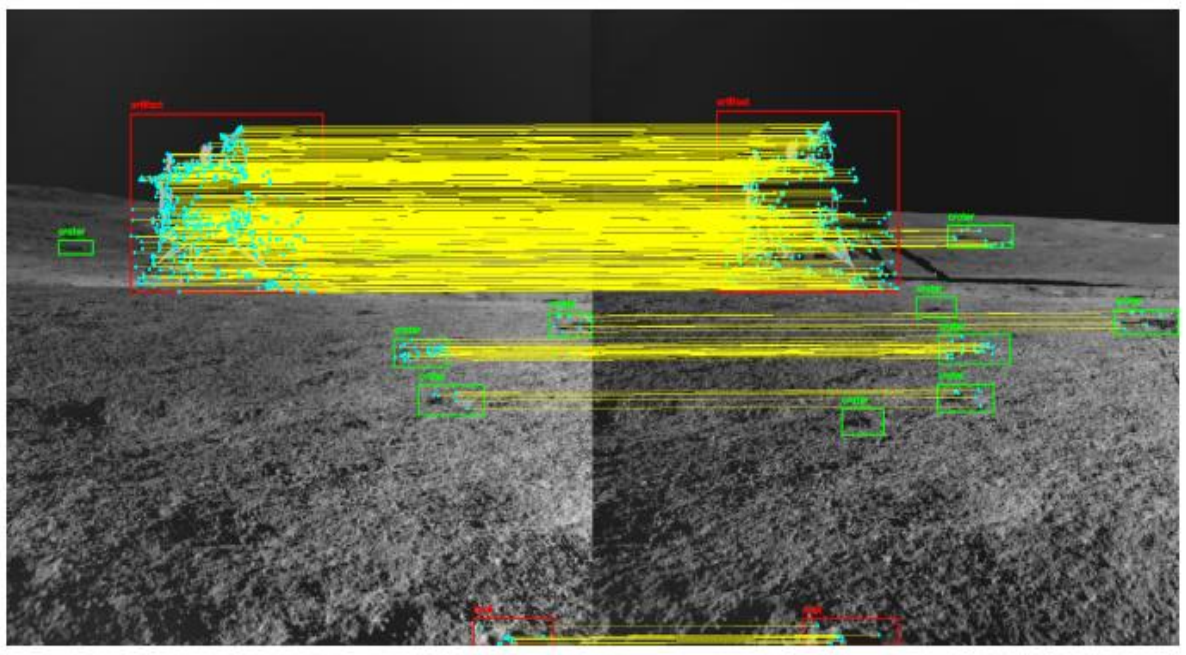

*Figure 3 Rocks, craters and artifacts detected in stereo images*

YOLO11n was selected for object detection after a comparative analysis with YOLOv8 [13] and YOLOv12 [14], due to its lightweight architecture and fast inference—qualities essential for deployment on resource-constrained planetary rovers. The model was trained using an open-source dataset of Apollo mission images [15], comprising 1,579 RGB images annotated with bounding boxes in COCO, YOLO, and Pascal VOC formats. These images were converted to grayscale and labeled for three classes: craters, rocks, and artifacts. An input resolution of 480×480 was chosen, offering an optimal balance between precision and recall for monochrome imagery, while reducing inference time and memory usage compared to the standard 640×640 setting. Most training hyperparameters followed Ultralytics' defaults, with adjustments including a batch size of 16, an initial learning rate of 0.001, and the use of aggressive data augmentations to improve generalization. Training was conducted over 150 epochs with early stopping based on validation loss. Model generalization was evaluated using stereo image pairs from the Chandrayaan-3 Pragyan rover [16], verifying performance on unseen lunar terrain.

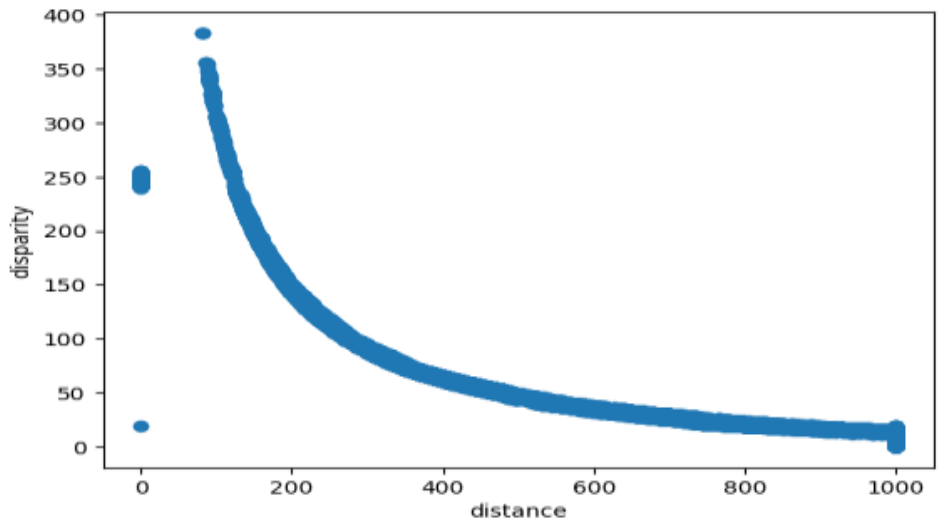

*Figure 4a. Inverse Proportionality of disparity and distances*

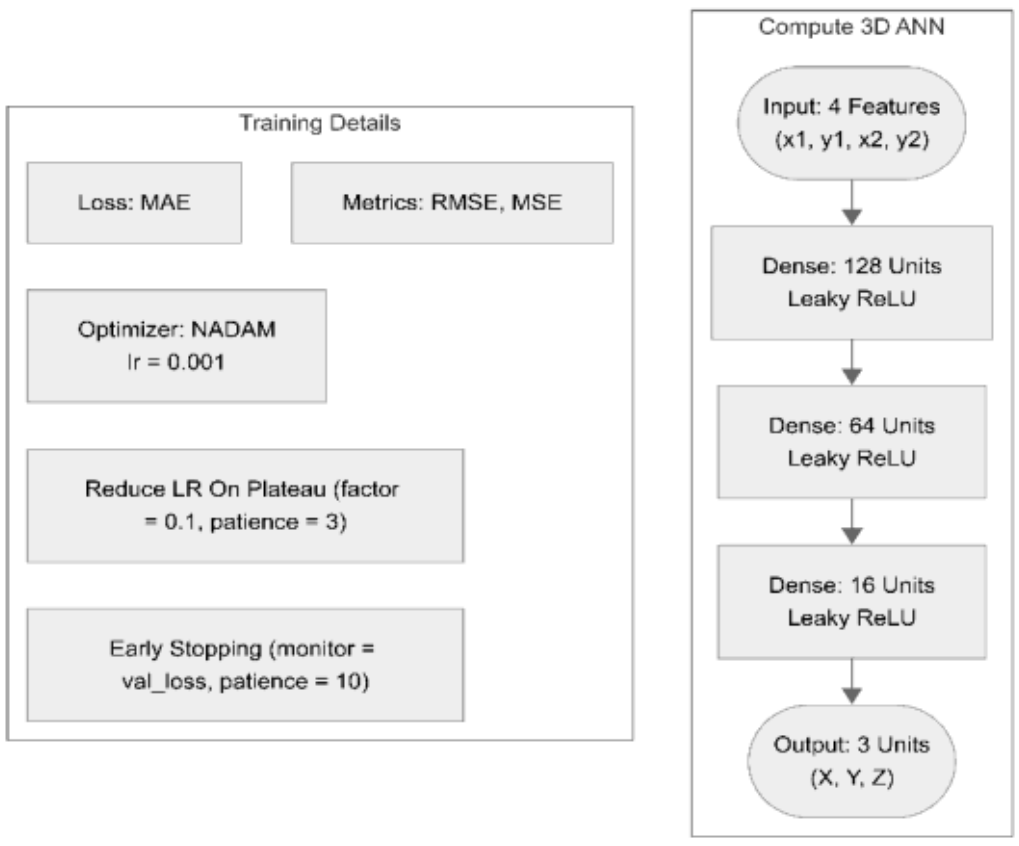

*Figure 4b. Artificial Neural Network for triangulation*

The regions within detected bounding boxes from both left and right images are designated as regions of interest (ROIs). SIFT features are extracted only within these ROIs, significantly reducing computational overhead and improving feature relevance. A two-way brute-force matcher is applied to match SIFT features bidirectionally across the stereo pair. This helps ensure accurate identification of corresponding objects—e.g., correctly associating the same rock detected in both images. Objects detected outside the region of stereo overlap or those lacking sufficient matches (fewer than four) are discarded. For each set of valid matched features, an artificial neural network (ANN) is used to triangulate and estimate the 3D coordinates (X, Y, Z). The ANN takes as input four values ($x_1$, $y_1$, $x_2$, $y_2$) and outputs the corresponding real-world coordinates. Training data was generated using ISRO's CAHV-based triangulation utility on real stereo pairs. A simple analysis of this data revealed that the distance value is inversely proportional to the difference in x pixel coordinates of a feature in the stereo image pairs, as shown in Figure 3a. The ANN architecture (figure 3b) includes three hidden layers with 128, 64, and 16 neurons respectively, each using Leaky ReLU activation, and is trained using Mean Absolute Error (MAE) loss with the NAdam optimizer. An initial learning rate of 0.001 and early stopping with patience of 10 were used. This combination was able to achieve minimum loss values and generalization without overfitting, focusing on capturing the relation observed in Figure 3a. The ANN achieved a training MAE of 0.15 centimetres.

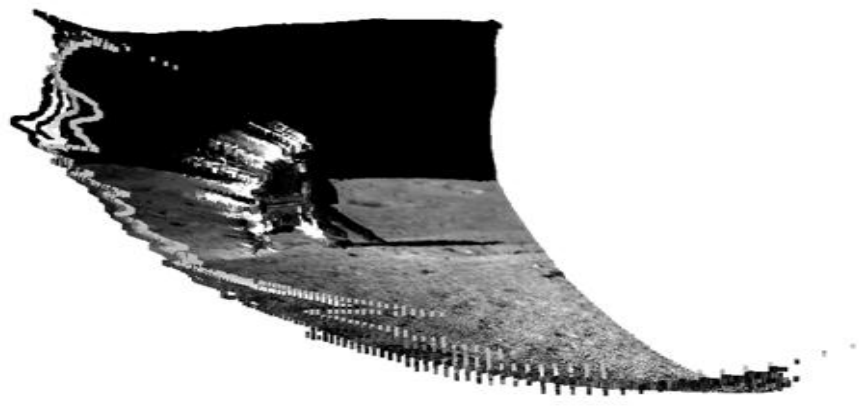

*Figure 5 Point cloud of Vikram Lander at Shiv Shakti Point, Moon*

For offline terrain reconstruction, the left grayscale image is passed through Depth Anything V2 (metric variant), which outputs a relative depth map aligned with the input resolution. This map is used with Open3D to construct a dense 3D point cloud of the terrain. The fusion leverages both the per-object metric distances estimated by the ANN and the dense depth map obtained offline to create visually coherent and metrically consistent reconstructions. The final point cloud is color-coded by input image intensity for visualization.

## IV. RESULTS AND DISCUSSIONS

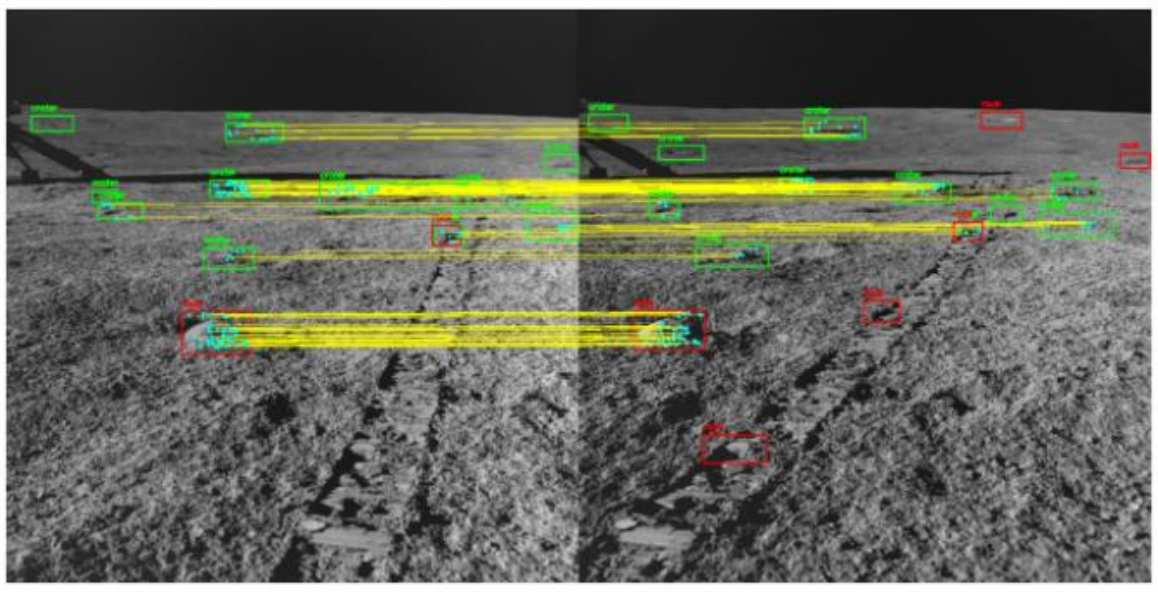

*Figure 6 Object Detection on Lunar Terrain*

A comprehensive comparison of YOLO model variants was conducted to evaluate detection accuracy, model size, and deployment suitability on grayscale lunar imagery. Among them, the YOLOv11n_imgsz640 configuration exhibited the highest precision while maintaining the smallest model footprint, making it ideal for applications where minimizing false positives and resource usage is critical. In contrast, YOLOv12s_imgsz640 achieved the highest recall (~0.64), prioritizing broader obstacle detection at the expense of increased computational cost and parameter count. Intermediate configurations such as YOLOv11s_imgsz480 offered a practical balance between recall and precision, making them suitable for general-purpose scenarios. An increase in input resolution consistently led to modest gains in detection accuracy, although with a significant rise in computational load. Across all tested families—YOLOv8, YOLOv11, and YOLOv12—the performance metrics were closely matched; however, YOLOv11 and YOLOv12 variants generally outperformed YOLOv8 in both accuracy and model compactness. Notably, YOLOv11n provided the highest precision, while YOLOv12n delivered better recall, showcasing their complementary strengths. Input resolution played a critical role: models with 480×480 inputs achieved substantially faster inference with minimal performance degradation compared to 640×640 variants. Overall, YOLOv11n_imgsz480 emerged as the most favourable choice, offering a strong balance between detection accuracy and computational efficiency—well suited for real-time perception tasks on planetary rovers.

| ID | Class | Predicted Distance (cm) - ANN | Distance (cm) - CAHV | Absolute Error (cm) |
|---|---|---|---|---|
| 1 | rock | 140.55 | 141.23 | 0.68 |
| 7 | rock | 150.18 | 154.64 | 4.46 |
| 6 | rock | 204.10 | 203.44 | 0.66 |
| 4 | rock | 214.63 | 213.87 | 0.76 |
| 2 | rock | 228.22 | 231.24 | 3.03 |
| 9 | rock | 336.99 | 340.86 | 3.87 |
| 0 | rock | 349.33 | 351.45 | 2.12 |
| 3 | artifact | 950.00 | 3915.83 | 2965.83 |

*Figure 7 Comparison of distance values derived from the image in figure 2 - ANN vs. CAHV tool results*

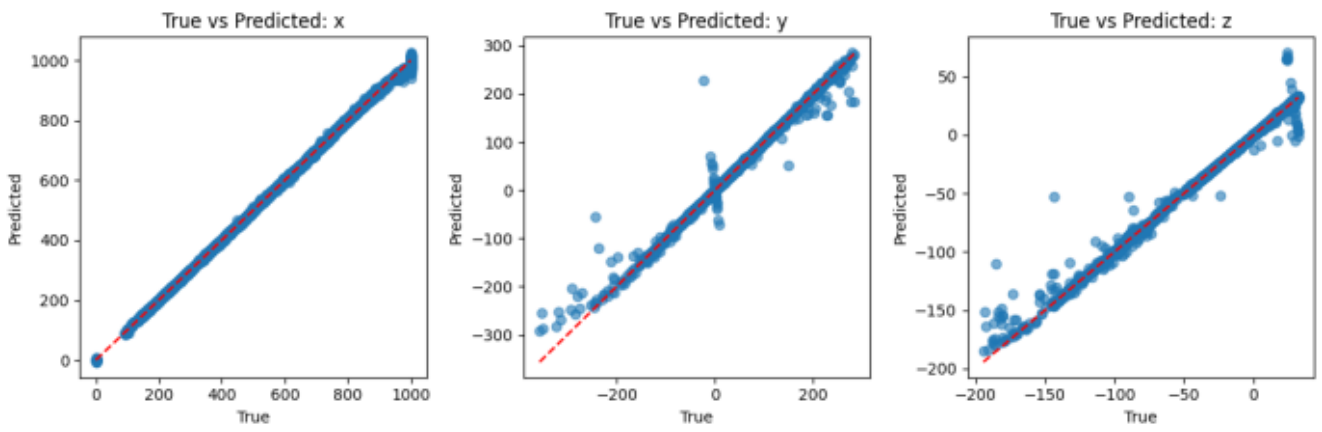

*Figure 8 True vs. predicted values of 3855 test samples*

The ANN-based triangulation method was evaluated against depth values generated by the standard CAHV-based geometric model. On a test set of 3,855 unseen samples (Figure 8), the ANN achieved a median absolute error of 2.26 cm, with an interquartile range of 0.91–5.58 cm—demonstrating reliable performance in the critical 1–10 meter range relevant for planetary obstacle detection. The model successfully preserved the inverse relationship between x-coordinate disparity and depth, ensuring geometric consistency with stereo vision principles. Unlike the CAHV pipeline, which performs sequential triangulation for each feature pair, the ANN enables scalable, batched inference through a single forward pass, offering significant computational efficiency. Notably, the ANN also delivered stable predictions for distant objects: beyond 10 meters, the model consistently capped depth estimates around this range, maintaining spatial coherence. In contrast, the CAHV tool often yielded inconsistent and unreliable depth values for different regions of the same object when its distance exceeded 10 meters, resulting in spatial inconsistency.
Furthermore, while the legacy CAHV utility required manual selection of matching features in stereo pairs, the proposed pipeline achieves full automation via YOLO-based object detection and SIFT-based keypoint matching. This automation streamlines the workflow, reduces manual effort, and minimizes user-induced errors.

## V. Conclusion and Future Work

The proposed vision system demonstrates an effective combination of lightweight real-time object detection and depth estimation with high-resolution offline terrain reconstruction, making it well-suited for planetary rover applications. The system achieves reliable obstacle detection and distance estimation under resource-constrained conditions. However, several components merit further refinement. The ANN-based triangulation model, while accurate, is purely data-driven. Incorporating physics-informed neural networks could embed geometric constraints and enhance generalization. Additionally, the current framework lacks autonomous control capabilities. Future work should explore integrating lightweight reinforcement learning (RL) policies for onboard decision-making. Finally, coupling the vision system with visual odometry, localization, and mapping frameworks tailored to lunar terrain would further strengthen spatial awareness and navigation robustness.

## Acknowledgment

The authors extend their sincere gratitude to the team members of Planetary and Space Data Processing Division (PSPD) for their invaluable vision, continuous support, and insightful guidance throughout this work.